# Short Term Prediction of Parking Area states Using Real Time Data and Machine Learning Techniques

**Jesper C. Provoost**
Department of Mathematics and Computer Science
University of Twente, Enschede, The Netherlands, P.O. Box 217, 7500 AE
Email: j.c.provoost@utwente.nl

**Luc J.J. Wismans**
Centre of Transport Studies
University of Twente, Enschede, The Netherlands, P.O. Box 217, 7500 AE
Email: l.j.j.wismans@utwente.nl

**Sander J. van der Drift**
Data scientist
DAT.Mobility
Deventer, The Netherlands, P.O. Box 161, 7400 AD
Email: svddrift@dat.nl

**Maurice van Keulen**
Department of Mathematics and Computer Science
University of Twente, Enschede, The Netherlands, P.O. Box 217, 7500 AE
Email: m.vankeulen@utwente.nl

**Andreas Kamilaris**
Department of Computer Science
University of Twente, Enschede, The Netherlands, P.O. Box 217, 7500 AE
Email: a.kamilaris@utwente.nl

Word Count: 6,210 words + 5 tables/figures = 7,460 words

*Submitted July 15th 2019*

**ABSTRACT**
Public road authorities and private mobility service providers need information on and derived from the current and predicted traffic states to act upon the daily urban system and its spatial and temporal dynamics. In this research, a real-time parking area state (occupancy, in- and out-flux) prediction model (up to 60 minutes ahead) has been developed using publicly available historic and real-time data sources. Based on a case study in a real-life scenario in the city of Arnhem, a Neural Network-based approach outperforms a Random Forrest-based one on all assessed performance measures, although the differences are small. Both are outperforming a naïve, seasonal random walk model. Although the performance degrades with increasing the prediction horizon, the model shows a performance gain of over 150% at a prediction horizon of 60 minutes compared with the naïve model. Furthermore, it is shown that predicting the in- and out-flux is a far more difficult task (i.e. performance gains of 30%), which needs more training data, not based exclusively on occupancy rate. However, the performance of predicting in- and out-flux is less sensitive for the prediction horizon. In addition, it is shown that real-time information of current occupancy rate is the independent variable with the highest contribution to the performance, although time, traffic flow and weather variables also deliver a significant contribution. During real-time deployment, the model performs 3 times better than the naïve model on average. As a result, it can provide valuable information for proactive traffic management as well as mobility service providers.

**Keywords:** Smart Parking, Machine Learning, Public data, Real-Time Prediction,

**INTRODUCTION**

Off-street facilities (e.g. parking garages) are usually distributed sparsely across a city and therefore require drivers to search more proactively for a suitable parking location [1]. Especially when a driver is unfamiliar of the area, or when traffic is heavy, this process wastes time and fuel while induces additional traffic load on the surrounding road network [2]. Searching for a vacant parking space thus imposes a significant burden on drivers and the wider economy, as valuable resources are wasted in the process. Considering that the number of passenger cars still increases to date, due to ongoing population growth, vibrant economies and urbanization, it is likely that these problems will increase as well. According to research [3], U.S. drivers spend an average of 17 hours searching for a parking spot every year. This amount is even higher in the U.K. and Germany with 44 and 41 hours per year, respectively. In Germany alone, the average driver wastes €896 per year on the hunt for a parking space. This aggregates to a yearly burden of €40.4 billion on the German economy. Furthermore, a survey of 17,968 drivers from 30 cities shows that 64% of participants experience stress while trying to find parking [3].

Traffic management applies measures to influence the demand and capacity of the traffic network in time and space to improve traffic operations [4]. The advance of modern technologies, particularly in the form of intelligent transportation systems (ITS), has supported authorities to execute their traffic management tasks more effectively and efficiently [5]. Within this context, many applications of ITS target the management of traffic by means of controlling infrastructure and access, e.g. by using lane management and signal control. However, ITS is also used as a means to directly inform or influence road users such that they make 'smarter' use of traffic networks [5]. With regard to parking, a relevant example includes the parking guidance and information (PGI) systems which supply drivers with dynamic parking information within controlled areas using road side equipment as well as in car information services [4,6].

What these ITS applications have in common, is their dependence on adequate and high-quality information, especially when non-regular traffic conditions occur [7]. Until now, authorities have mostly depended on real-time or historic data for these purposes. However, due to the highly dynamic nature of traffic, current information may become obsolete within a matter of minutes. This, combined with prevailing latency in data availability, limits the effectiveness of contemporary traffic management measures. Stakeholders, like public road authorities and private mobility service providers need information derived from the current and predicted traffic states to act upon the daily urban system and its spatial and temporal dynamics [7]. Accurate parking predictions may lead to better management of the system by transport operators and to a potential congestion mitigation due to avoidance of queue formation [8]. Predictions could be used as instrument to timely inform drivers, such that the effectiveness of their decisions is maximized upon arrival at their destination. Moreover, since 40% of traffic in urban areas is attributed to the search for a parking space [9], knowledge on traffic flows associated with parking (in- and outflux) provides valuable information for ITS applications (e.g. induced traffic loads as a result of parking and as an input for short term traffic state predictions of urban networks).

Earlier research on real-time predicting parking area states, has focused on occupancy rates (parking availability), using a variety of input data depending on availability [8,10-23]. Although this is an important variable to feed ITS systems, also the in- and out-flux of these areas provide valuable information, which is an important aspect included in this research. Furthermore, this paper focuses on the use of publicly open-access data and adds to the existing research by comparing machine learning techniques, analyzing their performance in a real-time application, investigating the data relevance and needs related to accuracy, as well as using innovative performance measures and analysis in a real-life case in the city of Arnhem, the Netherlands. This research shows that a Feed-Foward Neural Network (FFNN) outperforms a Random Forrest (RF) on all assessed performance measures, although the differences are small and both are outperforming a naïve models. Furthermore, it is shown that predicting the in- and out-flux is a far more difficult task which needs different training data, beyond occupancy rate. However, the performance of predicting in- and outflux is less sensitive for prediction horizon. In addition, it is shown that real-time information of current occupancy rate is the independent variable with the highest contribution to the performance of the machine learning models.

**BACKGROUND**
In this section, we will further discuss previous research on this subject focusing on the variables shown to be relevant for predicting parking area states and the used techniques. This knowledge is used for our research design.

**Relevant variables**
According to Guyon and Elisseeff [24], the predictive power of a model is highly dependent on the chosen variables. Feature selection is therefore a crucial task, not only to optimize performance, but also to provide a better understanding of the underlying processes. A selection of eleven articles was made, in order to determine the most promising predictive variables. Note that all selected articles solely consider the occupancy rate as dependent variable in their research.

Parking flows are dynamic over time, and therefore temporal variables are among the most prominent candidates in terms of predictive ability. Chen et al. [10] demonstrate that seasonal variables, such as *time and date*, lead to dramatically improved prediction accuracy. This is supported by Badii, Nesi and Paoli [11], who regard time variables as the baseline for their model. As a matter of fact, the variable *time of day* is mentioned in almost every article. Lijbers claims that the *weekday* variable would also enhance the model's accuracy [16]. Most articles support this, even though Hampshire et al. [12] and Badii et al. [11] suggest that the actual importance of this variable is quite low.

**TABLE 1 Matrix of independent utilization**

| Article | Variable | | | | | | | |
|---|---|---|---|---|---|---|---|---|
| | Time of day | Weekday | Temperature | Rain | Holiday | Event | Traffic flow | Historic occupancy |
| Vlahogianni et al [8] | X | X | | | X | | | X |
| Badii et al. [11] | X | X | X | | | | X | X |
| Hampshire et al. [12] | X | X | | X | | X | | |
| Chen [13] | X | X | | | | X | | |
| Zheng et al. [14] | X | X | | | | | | X |
| Camero et al. [15] | X | X | | | | | | |
| Chen et al. [10] | X | | | | X | | | |
| Lijbers [16] | X | X | X | X | X | | | |
| Monteiro and Ioannou [17] | | X | | | | | | X |
| Reinstadler et al. [18] | X | | X | X | X | X | | |
| Pflugler et al. [19] | X | X | X | | X | X | X | |

Additionally, *historic occupancy* is also regarded to be a strong predictor. Vlahogianni et al. [8] demonstrate using genetic optimization that a lookback time window of 5 minutes in the past may be efficiently used to predict parking occupancy (%) up to 30 steps in the future with high accuracy. Similarly, Zheng et al. [14] argue that a 30% performance gain can be achieved by including several steps from the past, in addition to just the time of day and weekday variables. Badii et al. [11], as well as Monteiro and Ioannou [17], suggest a similar effect. On the contrary, some of the other articles do not endorse the historic occupancy as input variable, which might be caused by the lack of availability of this data. For instance, Reinstadler et al. [18] define their research as a 'data mining problem', which entails that their data points are independent and unordered over time, unlike time series data.

In the majority of relevant articles, a weather variable such as *temperature* or *rain* is used as input of the model. Reinstadler et al. [18] argue that, because weather data has a high weight in their resulting model, these variables are very important for the accuracy of predictions. Nevertheless, Chen et al. [10] challenge this by stating that weather conditions (in a Dublin City case) showed little impact on parking occupancy. However, Badii et al. [11] demonstrate that the importance of temperature and rainfall varies significantly per distinct parking location which could explain these results.

The variables *event*, *holiday* and *traffic intensity* could supposedly provide a useful addition to the model, even though Pflugler et al. [19] claim that they are of secondary importance. This claim is supported by Badii et al. [11], who remark that the traffic flow variable is only relevant when sensors are located on streets leading to the parking garage, and when measurement data is available for the previous hour with respect to the time of prediction. Moreover, *event* and *holiday* might be important variables, since some articles mention these variables as important. For instance, Reinstadler et al. [18] state that external attributes like events and holidays are extremely important since they influence parking occupancy. Chen et al. [10] support this by demonstrating how the predictive error and standard deviation spike during the Christmas holidays.

Earlier research shows that time variables are the most prominent predictors for a machine learning model on parking occupancy. The historic occupancy, provided that a lookback window is possible, is shown to be an important predictor. Secondary to this, the variables *temperature, rain, holiday and event* could increase predictive power because of their supposed relationship with traffic flows, and consequently also the in- and outflux, even though this seems to be mainly related to a lack of research and reliable data sources. Hence, there is still a clear opportunity for these variables to be successfully applied onto the predictive model.

**Analysis of contemporary techniques**
Earlier research on predicting parking area states focuses on supervised regression, targeting occupancy rates. Stolfi et al. [20] compared six relatively simple predictive techniques on parking data from the city of Birmingham and observed that polynomial regression and time series prediction provide the best results. Zhu et al. [23] and Camero et al. [15] acknowledge this, but remark that there are more sophisticated techniques which can help to enhance the predictive accuracy. Reinstadler et al. [18] showed that regression trees generate better predictions than the ones mentioned by Stolfi et al. The authors argue that regression trees are more flexible and often more powerful than time series-based techniques, because the latter ones only consider the temporal seasonality patterns and cannot cover all predictor variables. Hampshire et al. [12] also concludes that the performance of the regression tree is superior compared to linear regression and time series techniques which might be caused by the fact that these techniques assume that all features are independent. A regression tree is able to expand the tree branches such that any correlation can be handled properly.

Additionally, ANN appear to produce promising results. Hampshire et al. [12] performed an analysis on two types of FFNN, and both proved to be more successful than a model based on linear regression. The use of ANN is further supported by Pflugler et al. [19] with the additional remark that ANN enable continuous learning in case that a real-time data feed is available. Yet, ANN are inconvenient due to their 'black-box' concept which prevents stakeholders from knowing the effect and influence of each variable. Furthermore, ANN could be unacceptable for real-time predictions due to their computational complexity. Research by Badii et al. [11] indeed confirm that training times are longer than regular regression methods. However, they also show that the actual time to make a prediction is only 0.0031 seconds, which is even less than the 0.0052 seconds it takes for a linear regression model. For real-time application, prediction times are far more meaningful than training times, predominantly since there is no need to retrain the model frequently. Recurrent neural networks (RNN) are a specific variant where nodes can form cycles and hence contain feedback loops. In this way, they are able to interpret sequences of inputs which rely on each other for context. For instance, the parking occupancy of one minute ago relies also on the occupancy of the occupancy two minutes ago, and so forth. Li et al. [22] demonstrate that LSTM (a specific kind of RNN) outperforms a regular FFNN on prediction of available parking spaces. The authors however remark that prediction times are significantly longer than traditional FFNN, which forms a bottleneck for a real-time predictive application.

In conclusion, regression trees are positively regarded by multiple authors because of their transparency as well as their ability to perceive correlations between variables. ANN have the potential to perform even better, even though they lack in their ability to provide transparent insights about the internal structure due to their black-box concept.

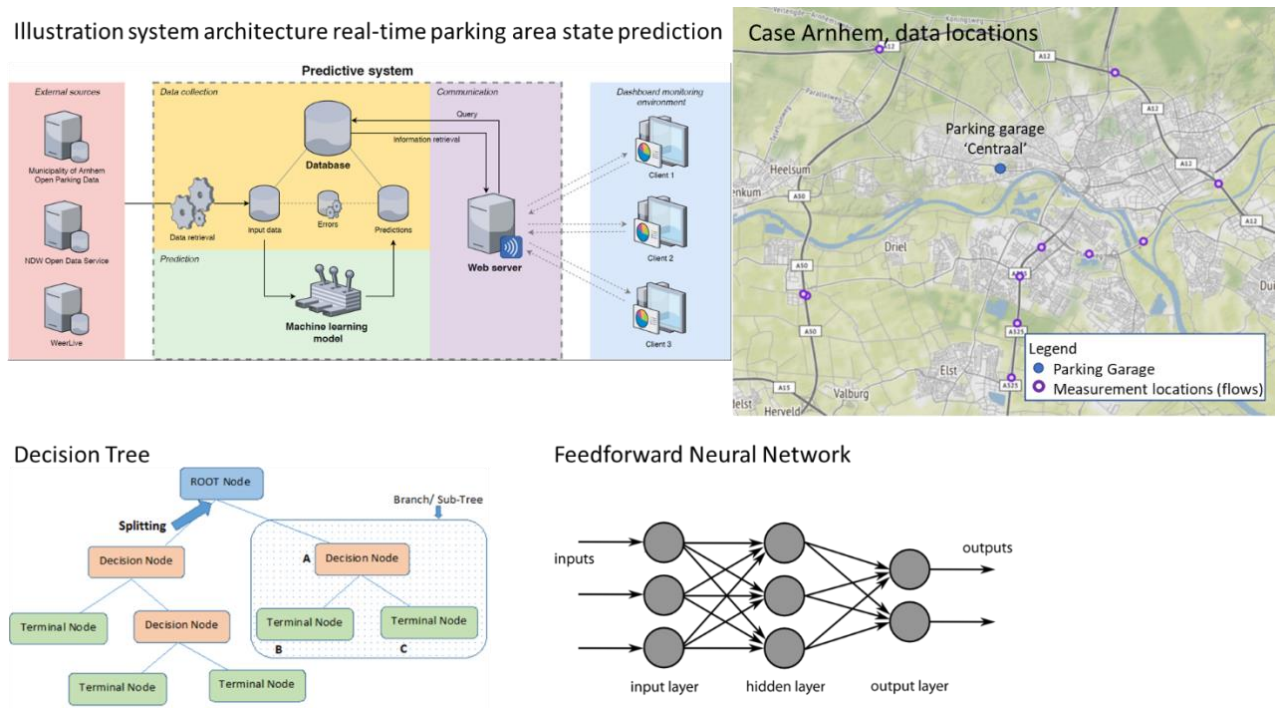

**Figure 1 Illustrations system architecture, machine learning techniques and case location**

**METHODS**

This section describes the selected machine learning techniques, the data collection and preparation, model development and the analysis and assessment framework used. The end goal is the real-time predicton of parking area states in terms of occupancy, in- and out-flux, which can directly be used for traffic management by mobility service providers (e.g. traffic information) or as input for larger traffic area state prediction systems [7] (i.e. parking zones are important origin and destination zones within urban areas). An illustration of this real-time system architecture is shown in Figure 1.

**Machine learning techniques**
Based on the literature presented in Section "Background", regression trees (RF) and the FFNN have been selected and are further explained below.

*Random forest*
Decision trees, which are generally applied to classification problems, utilize a tree structure to recursively classify input variables to a fixed set of output variables [25]. Upon training a decision tree model, the dataset is split into smaller and smaller subsets while an associated tree structure is incrementally built at the same time. Regression trees are a variant of conventional decision trees, with the obvious difference of being applicable to regression problems. Instead of classifying an outcome to a predefined set of categorical variables, regression trees output a numerical continuous value, e.g. the influx, outflux or occupancy rate of a parking area. When training a regression tree, every input variable is recursively partitioned based on minimization of the error between the predicted value and the actual value in the training set. New data can be filtered and lands into one of the leaf nodes which corresponds to a numerical value. This makes it possible to generate predictions.

Decision trees are known to suffer from bias and variance. Ensemble methods combine multiple trees in pursuance of increased robustness and better predictive performance. They are implemented in the form of bagging and boosting, which both produce new subsets of the training data by random sampling with replacement. Subsequently, each collection of subset data is used to train their respective decision trees, which results in an ensemble of models. Bagging techniques are used to make the resulting model less prone to individual trees overfitting the training data. A widely used implementation is RF, which takes one extra step as opposed to regular bagging techniques: in addition to randomly selecting subsets of data, it also takes the random selection of features to grow trees. Its prediction is given based on the aggregation of predictions from all trees in the model. The main advantage of RF is the potentially high performance while maintaining relative ease of implementation, especially since the tuning of hyperparameters is fairly easy. Generally speaking, finding the optimal balance between the number of trees in the model and decent computational performance is the most important aspect of hyperparameter tuning. Above all, RF generally provide good scalability and suitability to a wide range of machine learning problems. [25,26]

*Artificial Neural networks*
An artificial neural network (ANN) is a computational model which is inspired by the way a human brain processes information. This technique has proved to be successful across many applications of machine learning, including regression problems [26]. The fundamental unit in an ANN is a neuron, often called a node. It receives an input from one or multiple other neurons, or from an external data source. Each input has an associated weight, which is assigned based on its relative importance to other inputs. Subsequently, in order to produce an output value, an activation function is applied to the given inputs. Additionally, a bias input contributes a constant value to the function, which may be critical for successful learning. Frequently used activation functions are ReLU, Softmax, Sigmoid and Tanh.

The feed-forward neural network (FFNN) is the conventional type of ANN. It contains multiple neurons which are arranged in layers. Neurons from adjacent layers have connections between them (each with an associated weight), such that the outputs from one layer of neurons serve as inputs for the next layer. FFNNs are very useful to overcome the problem of non-linearity. In combination with their flexible structure, i.e. the ability of adding or removing neurons and hidden layers to the model, this makes them applicable and scalable to a wide range of tasks. By the same token, the output layer can contain an arbitrary number of neurons, which makes this technique suitable for multi-output predictions [26]. This is particularly useful when predicting time series, where each predictive horizon (i.e. 1 minute ahead, 5 minutes ahead, 10 minutes ahead and so on) can be represented by its own output node, which makes it theoretically suitable to the task of predicting flows and occupancy rates of parking areas on a horizon of up to 60 minutes.

**Data collection and preparation**
In order to develop and operate a functional predictive model, both historical and real-time data sources should be available and operational. Based on the literature, the initial variables were selected for which historical data sources are needed as the basis to develop and tune the model (i.e. training, validating and testing). This historical dataset comprises of a vast number of entries which contain a value for each independent and dependent variable. Subsequently, to actually make predictions with the resulting model, real-time data sources should be accessible in order to provide actual values to the input of the model and as a result provide the constraints applied when preparing the dataset for developing the model.

Parking data is inevitably the most crucial data source within this research, given the fact that the intended model aims to predict the three parking variables occupancy rate, influx and outflux. Although historical open data sources for parking areas are still scarce today, the Open Parkeerdata portal of the Municipality of Arnhem [27] provides historical transaction cost data which is used for deriving the historical occupancy rate as well as the in- and outflux. This portal also provides a real-time data feed providing updates of the occupancy rate, approximately every 11 minutes, which resulted in the constraint that real-time in- and outflux were not part of the independent variables taken into consideration. The data source provides transaction data of three parking garages in Arnhem of which the 'Centraal Garage' has been chosen as being the largest one in the city centre. Data was used from August 2017-April 2019.

Traffic data (flows in veh/h for every minute) was gathered from the Nationale Databank Wegverkeersgegevens (NDW) using its Dexter [28] platform (historical data) as well as the Open Data Service of NDW providing the real-time feed. In total, eleven measurement locations were selected, all of which are part of the MoniCa loop detection system operated by Rijkswaterstaat, which means they are situated on the orbital highways and freeways around Arnhem, specifically on highway exits and access roads. After considering the availability and validity of the measurement sensors, Traffic flow data was used from November 2017-April 2019. Since traffic flows are sensitive to randomness and high variance, smoothing was applied. For this purpose, a 2nd order low-pass Butterworth filter (with a cutoff frequency of 0.05) was applied. This method was selected in favour of a regular rolling mean, mainly since the rolling mean introduces a lag which will be problematic when real-time data sources are used.

The open databases of the Dutch meteorological institute KNMI [29] were utilized for the historical weather data and the Weerlive Api (based on KNMI data) for the real-time feed. Using a web service, the hourly data of several weather-related variables can be queried. The measurements of the closest weather station were chosen (i.e. Deelen station 10 km from the city center. The data source provides the air temperature at 1.5-meter height (measured in 0.1 °C) and rainfall (a binary variable denoting whether rain has fallen in the past hour) variables at a 10-minute interval. The hourly data from August 2017-April 2019 were used.

All data was checked on plausibility and cleaned. When preparing the complete dataset with cases a complete case was deleted if data of a certain independent variable was missing for that specific time instance. Because there was limited missing data, the data cleaning resulted in less then 0.1% deletion of cases. All data was resampled on a one-minute interval as well as a lookback window was added for occupancy (the past 5 real-time updates (every 11 minutes), which means approximately 1-hour lookback window) and flow (the past 3, 10 minute rolling sum intervals, which means approximately 30 minutes lookback window). In the end a total of 544,680 cases were available.

The prediction intervals were set to predict up to 90 minutes ahead for every 5-minute interval. Since the real-time occupancy of the Centraal garage is disclosed every 11 minutes, the 30-minute buffer assures that the real-time predictive system is always able to predict for 60 minutes ahead - even in the worst case where the last occupancy rate was received 10 minutes ago.

**Model development**
The overall process of model development is divided into three phases:
1. During model architecture selection, a systematic search is applied to establish the optimal internal architecture for both model types. Based on [25] this concerned the number of neurons and number of hidden layers for the FFNN and the number of trees for the RF. This phase results in two preliminary models.
2. Both models are then optimized during hyperparameter tuning. Relevant parameters of both model types are systematically tweaked, followed by repeated evaluation of the model performance on the validation set. Based on [25] this concerned the learning rate for the FFNN and for the RF the maximum tree depth and maximum features. This results in two candidate models (i.e. a FFNN and a RF).
3. In the inter-model comparative testing phase, the candidate models of both types are assessed using the test set against a naive model. The methodology behind this phase is further described in the next paragraph on the assessment framework.
4. The best performing model was tested while applying it in real-time mode against a naïve model.

To systematically find the optimal configuration for both model types, a grid search was applied in phase 1 and 2 by defining subsets of the relevant parameter spaces. Subsequently, all possible combinations of those subsets were tested by compiling, training and validating a new model to find the best combination of parameter configurations using the performance measures of the assessment framework.

To be able to train and test the methods, the complete dataset is split into three datasets (training, validation and test). The training set is used by the model to learn the actual patterns, the validation set is used to understand the behavior of the preliminary model and its generalizability (avoiding under- or overfitting during the hyperparameter tuning by evaluating the loss on the validation set). The test set is kept separate until the very end, used to test the performance of the resulting models. Usually the validation and test set are kept small compared to the training set [13], in this research we used 72% training, 8% validation and 20% test. Given the sequential nature of the input data (i.e. time series), we maintained the chronological order of data, such that the model's sensitivity to seasonal patterns will become more evident during the validation and testing phase. This means that the selection of cases being part of the various sets has not been done by random selection.

**Assessment framework**
*Metrics*
Regarding the actual performance assessment, a combination of mean squared error (MSE), mean absolute error (MAE) and mean absolute scaled error (MASE) is used. MSE and MAE are often used and provide a comprehensible and precise way of understanding the magnitude and distribution of the model's errors using a natural, unambiguous scale. These measures are used for the architecture selection and hyperparameter tuning. However, these metrics have the disadvantage of becoming unifity or undefined when the occupancy or in- and outflux approaches zero. MASE, earlier used by [11], compares the model's MAE to that of a naive benchmark model, which makes it robust to scaling differences. Moreover, it demonstrates the added value of each model compared to a naive model, providing additional insights in performance.

$$MASE = \frac{MAE}{MAE_{Naive}} \tag{1}$$

$$MAE = \frac{1}{n}\sum_{j=1}^{n}\left|y_j^{pred} - y_j^{actual}\right| \tag{2}$$

$$MSE = \frac{1}{n}\sum_{j=1}^{n}\left(y_j^{pred} - y_j^{actual}\right)^2 \tag{3}$$

*Naïve benchmark model*
In order to use MASE, a naive benchmark model must be defined first. Two commonly used naive models are the random walk and the seasonal random walk. The random walk uses the last known observation to predict the future values (i.e. the last known occupancy rate from the Centraal garage will be used as prediction for 5 minutes ahead, and also 90 minutes ahead). The seasonal random walk model incorporates seasonal and temporal patterns in order to make predictions (e.g. the influx from one year ago would be used to predict the influx for the upcoming minute). Because we are interested in the performances for various prediction horizons as well as the known similar daily/weekly patterns the seasonal random walk in which the known occupancy rate, in and outflux of one week ago is considered to be the most challenging comparison for the developed models for the various (especially larger) prediction horizons.

*Quality of predictions/analysis*

The presented metrics are used to determine the quality of the models. Next to overall performance on the test set, also the error distribution was considered using a violin plot to understand to what extent the model shows consistent prediction quality and to detect possible circumstances which were difficult to predict. A violin plot is similar to a conventional box plot, but has the additional benefit of displaying a rotated density plot on both sides. Furthermore, because the purpose is real-time prediction, the prediction time (i.e. the time needed to predict the dependent variable once the model was trained) using the real time data feeds, was measured.

The best performing model was also applied in real-time, running the model for one week, and evaluated using the MASE metric as well, analyzing the error magnitudes to understand to what extent a real application would result in similar performance than the off-line tests.

Next to the performance for the case Arnhem, it is also of interest to test whether the applied methodology can be transferred to other locations. For this purpose, we analysed the input variable dependency using a feature elimination strategy and the impact of limited training data. Feature eliminiation entails that variables are categorically removed from the input dataset. For every variable (or category of variables) that is removed, a model is trained with the remaining input columns. The performance of this model is then comparing the MSE on the test set of this new model with the original reference model. Since some input variables are interrelated (e.g. the lookback windows), these were categorized such that they could be eliminated collectively. During the test, all identified categories were then separately eliminated, such that only a single category of variables was absent during every test round. The impact of limited training data was tested by recursively dividing the training set into halves, and training a new in/outflux and occupancy rate model every time based on the resulting subset. The data was not shuffled to maintain the natural order of the time series. Hence, the oldest half is removed from the set, while the most recent half remains for the next round. In this test the MASE metric was used to determine to what extent the model was still capable of outperforming the naïve model.

## RESULTS
### Candidate models
For both models the candidate models are developed by optimizing the architecture and the hyperparameters using grid search. This section describes the outcomes and the performance of the resulting candidate model per methodology.

*FFNN*
Based on the grid search (see the heatmap, Figure 2), it is shown that the worst performance is found at the bottom right of the figure where a low number of neurons is divided over a large number of layers, which leads to underfitting the model. The best performing models are located around the top-left and middle-left areas of the heatmap. This suggests that the FFNN performs best with an architecture in which a high number of neurons is divided between a relatively small number of hidden layers. The minimum MSE was observed at configuration (90, 4), i.e. with 90 neurons spread across 4 hidden layers.

Next, the hyperparameter of learning rate was optimized. For all learning rates a, the progression of MSE over time (i.e. the number of epochs) was visualized using a line graph (see Figure 2). As expected, higher learning rates initially show a rapid decrease of MSE, but then stagnate and show unbalanced behavior, the lower learning rates, on the contrary, demonstrate a stable decrease but converge too slowly. According to the previously defined criteria, the learning rate *α = 0.0001* provides an optimal balance: after 200 epochs, the corresponding loss is the lowest and is still descending at a substantial pace.

Using these parameters, the final FFNN was trained over the course of 2000 epochs. To prevent overfitting, a model checkpoint was applied to save the model's parameters at the epoch where the lowest validation loss was measured.

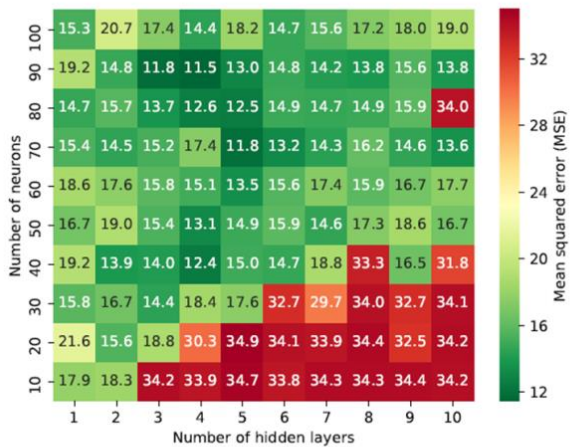
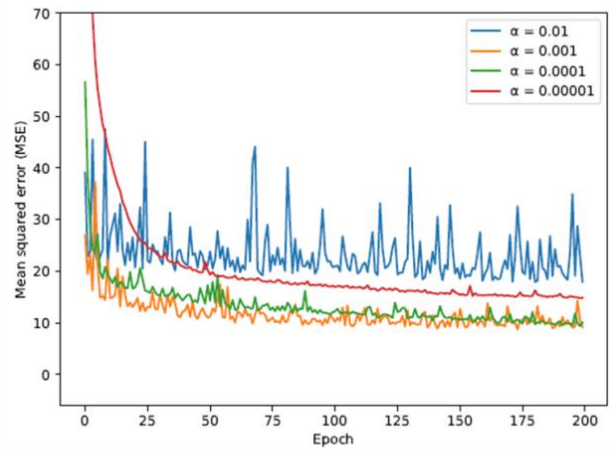
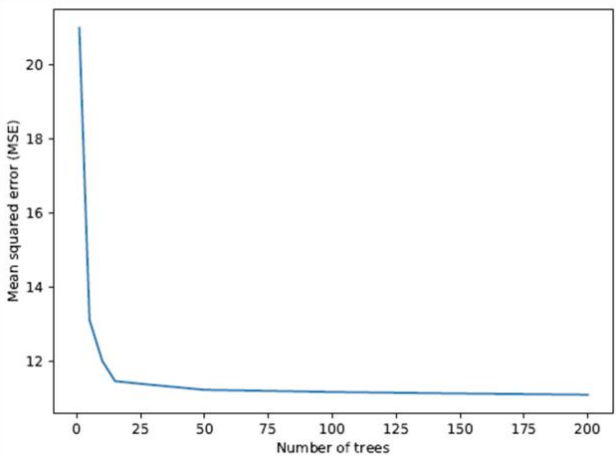
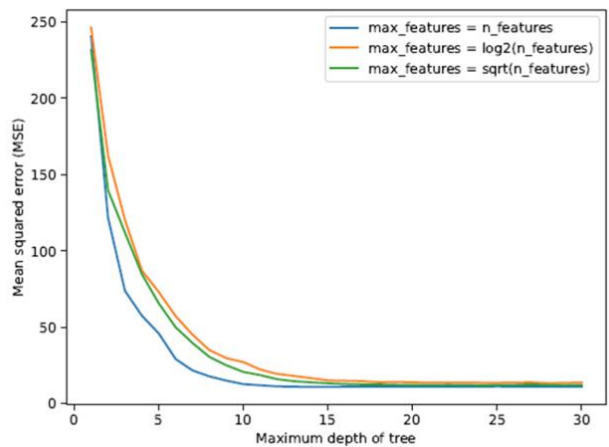

**Figure 2 Results tuning methodologies**

*RF*
The performance of the RF depends on the number of trees selected. The results suggest that the validation loss is subject to exponential decay when the number of trees *n* increases. When there is only one tree in the ensemble, the RF can essentially be regarded as an ordinary decision tree. The real power of the RF becomes evident when the number of trees grows. Around *n = 50*, the MSE seems to reach a plateau state. A higher number of trees would thus be ineffective: no significant performance gain will occur anymore, even though the computational complexity will rise dramatically.

Based on a grid search, the hyperparameters maximum features and maximum depth of the trees. From the results, it becomes clear that the three configurations follow the same trend in relation to the maximum depth *d*. Nonetheless, in the case where the maximum features equal the available number of features, the validation loss clearly decreases faster and reaches the plateau state at a significantly lower value of maximum depth. This is the preferred option, since the maximum depth should be rather small in order to minimize the computational complexity (i.e. training and prediction times). Using this configuration, the minimum MSE is reached at a maximum depth of 12, after which no further performance gain takes place anymore. Using these parameters, the final RF was trained.

**Comparison of models**
The final candidate models are compared using the MSE, MAE and MASE measure for the three predicted dependent variables (i.e. occupancy, in- and outflux) for the various prediction horizons as well as the error distribution (see Figure 3).

Even though the differences are not large, the results demonstrate that the feed-forward neural network outperforms the random forest in every aspect. The results also show that every model predicts significantly better than the naive benchmark. In particular, the occupancy rate prediction is exceptionally well: across all horizons (including the buffer), the corresponding MASE is around 0.40 which suggests that the performance gain is 150% compared to the naive model. The MAE is only 1.65% across all horizons and at a prediction horizon of 60 minutes 2.02% showing its high accuracy. Regarding the influx and outflux models, the overall MASE is between 0.75 and 0.8 and the MAE remains rather constant at 3.9 across all horizons. Note that the occupancy can vary between 0% and 100%, while the in- and out-flux varies between 0 and 97 vehicles. As the summarized overview of metrics already suggested, the MASE plots confirm that the FFNN outperforms the RF. However, the differences become smaller as soon as the predictive horizon increases. In fact, in case of the outflux, the RF starts to to outperform the FFNN after predicting approximately 65 minutes or further ahead. A similar phenomenon occurs at the influx, where the RF starts to perform better after 85 minutes.

Notably, predicting occupancy rates is more reliable than predicting the in- or out-flux and the prediction accuracy of the in- and out-flux seems to be less sensitive for the prediction horizon. This is probably related to the fact that no preceding in- or outflux information was used to estimate the model, which was the case for occupancy, but is probably also related to the lower variance in occupancy rate over time compared to in- and outflux (also see the transferability analysis below). Furthermore, the performance on influx is slightly higher than the outflux which is related to the larger expected correlation of influx with flows (and preceding flows).

The violin plots showing the error distributions show that the center of gravity is located under the MASE = 1 line, which indicates that all models predominantly predict better than the naïve model. Yet, the distributions of influx and outflux errors contain more irregularities than the distribution of occupancy rate error. Also, the distributions are more skewed towards a higher MASE (above 1), which makes it evident that a significant number of in- and outflux predictions are worse than those of the naive model. In contrast, almost none of the occupancy rate predictions are worse. Not surprisingly, all models (Naïve, RF and FFNN) have most difficulties in predicting correctly during specific non-regular conditions (e.g. Kingsday and large events). However, the average MAE for these specific days remains below 2.5% for FFNN and RF, while the naïve model shows MAE values of over 15%. The outliers in the MASE metric in which the naïve model performs better turn out to be associated with conditions in which the occupancy rates are near zero for which the MASE metric is sensitive.

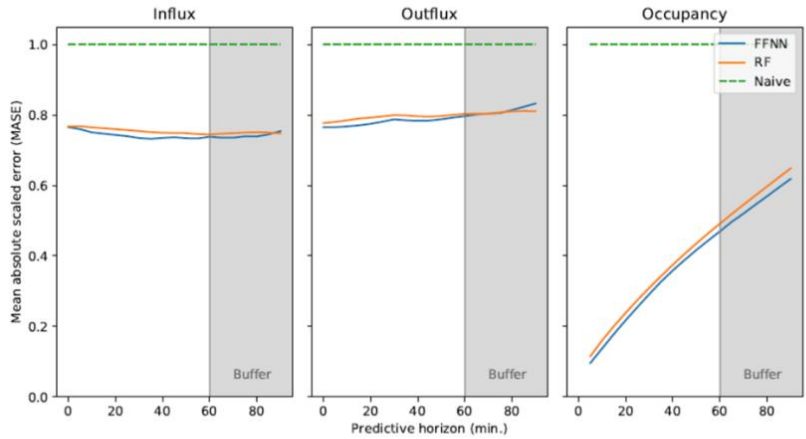

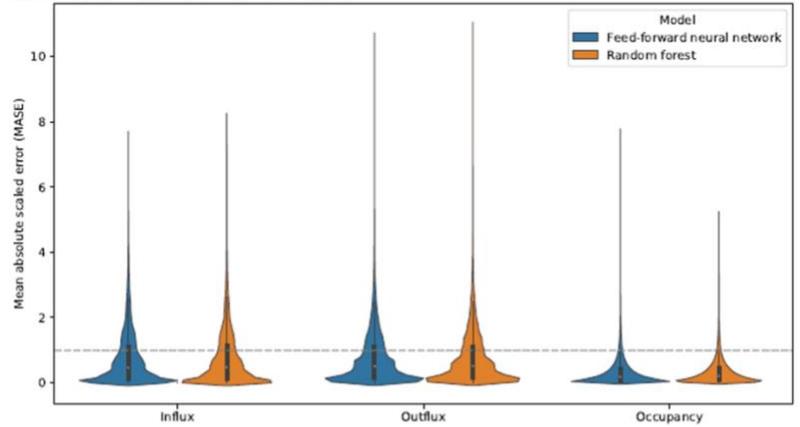

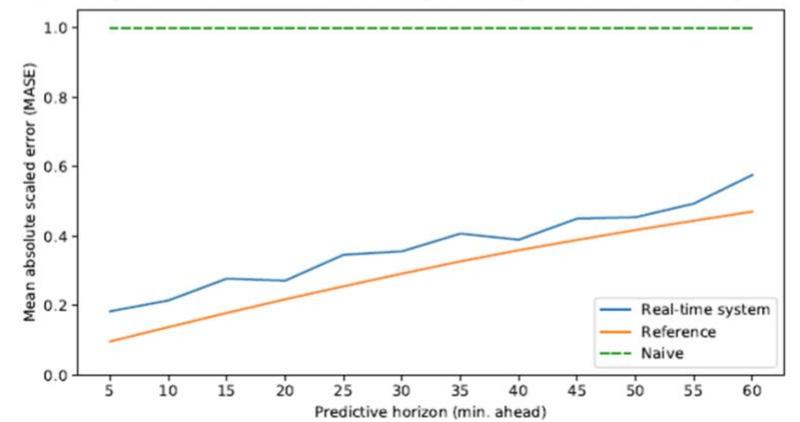

**Figure 3: Results comparison models**
When the model is deployed in real-time, the system turns out to predict worse than the standalone FFNN did on the historical test set. Notably, the MSE has doubled and the MASE has increased by almost 0.1. A likely cause for this difference would be the significant delay of incoming occupancy rate measurements (i.e. 11 minutes). In the worst case, the system does therefore not possess any information about the last 11 minutes, and will therefore have to use a large part of the buffer. This increases the uncertainty of predictions, and hence the magnitude of errors. This highlights the importance of a reliable and frequent input feed. It should be noted, however, that the current real-time system still delivers predictions with a very high quality: the average MASE is 0.37 across the prediction horizons 5-60 minutes, which amounts to a performance gain of 170% with regard to the naive model. The RF needs 1.32 seconds for a prediction which is slightly faster than the FFNN which needs 1.57 seconds.

**Transferability analysis**

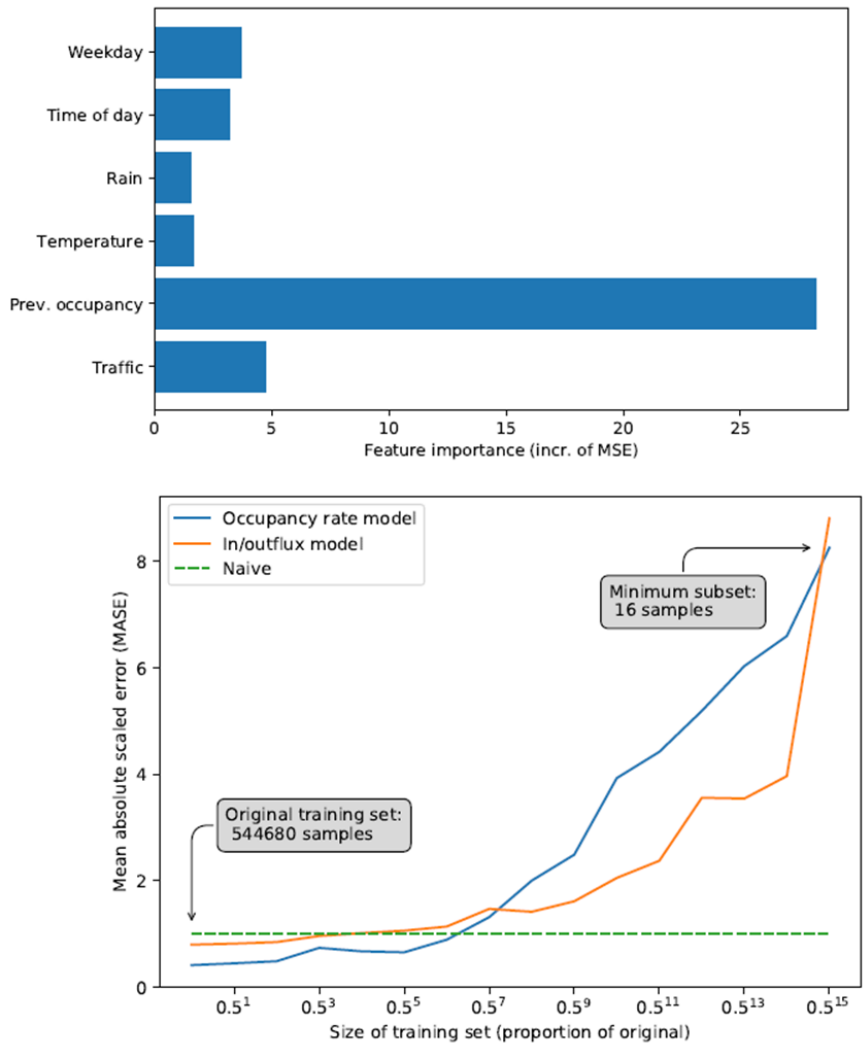

**Figure 4: Results transferability analysis**

Based on the overall performance, the FFNN has been selected to further analyse the input variable dependency and impact of limited training data. Both aspects? provide important information on the suitability of the approach to use for other locations. The results of the feature elimination strategy (see Figure 4) show that the preceding occupancy is by far the most important feature required by the model. However, it is also shown that the traffic flow and time variables do have a significant importance for the model (showing an increase of MSE of approximately 5 compared with the original MSE of 6.37) and in a lesser extent weather.

    The impact of data availability shows that for occupancy, up to halving the dataset 6 times (i.e. less than 6 previous days of data available) the FFNN model performs better than the naïve model. For the in- and outflux much more data is needed (i.e. approximately one-and-a-half month). Nonetheless this means that it is not necessarily needed to build large historic databases to be able to estimate a reasonable model, although the performance would benefit from this.

**CONCLUSIONS AND FURTHER RESEARCH**
Within this research, we compared two promising machine learning methods with several publicly available data sources, towards the real-time prediction of parking area state occupancy, in-flux and out-flux in a real-life case. The methods and independent variables were based on earlier research. The available historic and real-time data feeds served as a constraint for the methodology. This research shows that the Feed Foward Neural Network (FFNN) outperforms the Random Forrest (RF) on all assessed performance measures, although the differences are small and both are outperforming a naïve seasonal walk model. Furthermore, it is shown that predicting the in- and out-flux is a far more difficult task which needs more historic training data than occupancy rate. This probably relates to the differences in data variance and the lack of real-time datafeed for in- and out-flux. However, the performance of predicting in- and outflux turns out to be less sensitive for the prediction horizon. In addition, it is shown that real-time information of the current occupancy rate is the independent variable with the highest contribution to the performance of the machine learning models, with time variables and traffic flow variables having only a secondary importance. During a real-time deployment, the developed model provides far better predictions up to 60 minutes ahead than the naïve model based on the parking area states of a week ago. As a result, it can provide valuable information for proactive traffic management as well as mobility service providers.

    Further research will focus on extending the model (deployment) towards other parking areas, improving the in- and out-flux prediction, integrating the model in other traffic state prediction models to be able to provide complete predictions of urban traffic states. Due to the increasing availability of real-time urban traffic flows (e.g. via connected traffic light controlers) it is expected that it is possible to improve the in- and out-flux. Additionally, we would like to extent the models by adding a measure for reliability based on further analyzing the error patterns and possibly using these as well for improving the prediction method. Finally, the application of the model to support pro-active traffic management will be further researched to assess the actual value of predictions for traffic management purposes.
Perhaps try LSTM as a candidate model in the future?

**AUTHOR CONTRIBUTIONS**
The authors confirm contribution to the paper as follows: study conception and design: J.C. Provoost, L.J.J. Wismans, S.J. van der Drift, M. van Keulen and A. Kamilaris
data collection: J.C. Provoost; analysis and interpretation of results: J.C. Provoost

Author; draft manuscript preparation: L.J.J. Wismans.
All authors reviewed the results and approved the final version of the manuscript.